\pdfoutput=1

\documentclass[11pt]{article}

\usepackage{ACL2023}

\usepackage{times}
\usepackage{latexsym}

\usepackage[T1]{fontenc}

\usepackage[utf8]{inputenc}

\usepackage{microtype}

\usepackage{inconsolata}

\usepackage{amsfonts}
\usepackage{amsmath}
\usepackage{arydshln}
\usepackage{CJKutf8}
\usepackage{graphicx}
\usepackage{multirow}

\title{CSS: A Large-scale Cross-schema Chinese Text-to-SQL Medical Dataset}

\author{Hanchong Zhang$^{1*}$, Jieyu Li$^{1}$\thanks{\ \ The first two authors contributed equally to this work.} , Lu Chen$^{1\dagger}$, Ruisheng Cao$^{1}$, \\
\textbf{Yunyan Zhang$^{2}$, Yu Huang$^{2}$, Yefeng Zheng$^{2}$ and Kai Yu$^{1}$}\thanks{\ \ The corresponding authors are Lu Chen and Kai Yu.} \\
$^{1}$X-LANCE Lab, Department of Computer Science and Engineering \\
MoE Key Lab of Artificial Intelligence, SJTU AI Institute \\
Shanghai Jiao Tong University, Shanghai, China \\
$^{2}$Tencent Jarvis Lab, Shenzhen, China \\
{\tt \{zhanghanchong,oracion,chenlusz,kai.yu\}@sjtu.edu.cn}}

\begin{document}
\maketitle
\begin{abstract}
The cross-domain text-to-SQL task aims to build a system that can parse user questions into SQL on complete unseen databases, and the single-domain text-to-SQL task evaluates the performance on identical databases. Both of these setups confront unavoidable difficulties in real-world applications. To this end, we introduce the cross-schema text-to-SQL task, where the databases of evaluation data are different from that in the training data but come from the same domain. Furthermore, we present CSS\footnote{Our code is publicly available at \url{https://github.com/X-LANCE/medical-dataset}}, a large-scale \textbf{C}ros\textbf{S}-\textbf{S}chema Chinese text-to-SQL dataset, to carry on corresponding studies. CSS originally consisted of 4,340 question/SQL pairs across 2 databases. In order to generalize models to different medical systems, we extend CSS and create 19 new databases along with 29,280 corresponding dataset examples. Moreover, CSS is also a large corpus for single-domain Chinese text-to-SQL studies. We present the data collection approach and a series of analyses of the data statistics. To show the potential and usefulness of CSS, benchmarking baselines have been conducted and reported. Our dataset is publicly available at \url{https://huggingface.co/datasets/zhanghanchong/css}.
\end{abstract}

\section{Introduction}

Given the database, the text-to-SQL task~\citep{zhongSeq2SQL2017, xu2017sqlnet} aims to convert the natural language question into the corresponding SQL to complete complicated querying. As the wild usage of relational database, this task attract great attention and has been widely studied in both academic and industrial communities.



Recently, text-to-SQL researches~\cite{hui-etal-2022-s2sql,lin-etal-2020-bridging,qi-etal-2022-rasat} mainly focus on building a parser under a cross-domain setup~\cite{yu-etal-2018-spider,wang-etal-2020-dusql}, where the databases of the training set and the evaluation set do not overlap. It aims to construct a universal parser that can automatically adapt different domains to inhibit the problem of data scarcity. However, domain-specific knowledge, especially domain convention, is crucial but difficult to transform across different domains under cross-domain setup. Another line of research focuses on the experiment environment where the training data and the evaluation data are based on the same database, which is known as a single-domain setup. A single-domain text-to-SQL system can parse domain knowledge more easily and also has more wide applications in the real world. However, the problem of data scarcity always comes up when security issues and privacy issues exist. Therefore, both of these setups will face particular difficulties when it comes to the real world.

To this end, we introduce the cross-schema setup in this work. The cross-schema text-to-SQL tasks aim to build a text-to-SQL parser that can automatically adapt different databases from the same domain, which can avoid the aforementioned problems. Actually, the cross-schema text-to-SQL also has broad applications in the real world. For example, all the hospital store the information of patients and medical resources in databases with different structures. Most information categories are identical across these databases, for instance, the patient name and the treatment date. Moreover, domain-specific representations such as medicine names in databases and user questions are also commonly used. In this case, we can build a universal in-domain text-to-SQL parser that can be deployed on the new database from the given domain. Compared with the cross-domain setup, a cross-schema parser will not always confront completely unseen domain knowledge. On the other hand, compared with the single-domain setup, the problem of data scarcity can also be inhibited because the data from other in-domain databases can be used to train the model. However, a cross-schema text-to-SQL parser need to automatically adapt different database schema structure. Unfortunately, this issue is less investigated before. Therefore, how to construct a structural-general parser is the mainly challenge of cross-domain text-to-SQL.

In this paper, we propose a large-scale \textbf{C}ros\textbf{S}-\textbf{S}chema Chinese text-to-SQL dataset~(CSS), containing 33,620 question/SQL pairs across 21 databases. We generate (question, SQL) pairs with templates and manually paraphrase the question by crowd-sourced. For the databases, we collect 2 real-world database schemas involving medical insurance and medical treatment. As the privacy issues, we are not allowed to use the original data. Therefore, we fill the databases with pseudo values. Based on these 2 seed databases, we alter the schema and expand 19 databases with different structures. Hence, CSS can be used to develop cross-schema text-to-SQL systems. On the other hand, the original 2 databases correspond 4,340 samples, which construct the largest Chinese single-domain corpus. This corpus also allows researchers to carry on related studies. Our main contributions can be summarized as follows:

\begin{enumerate}
    \item We present the cross-schema text-to-SQL task and propose a large-scale dataset, CSS, for corresponding studies. The dataset and baseline models will be available if accepted.
    \item We provide a real-world Chinese corpus for single-domain text-to-SQL researches.
    \item To show the potential and usefulness of CSS, we conducted and reported the baselines of cross-schema text-to-SQL and Chinese single-domain text-to-SQL.
\end{enumerate}

\section{Related Works}

\paragraph{Single-domain text-to-SQL datasets} Earliest semantic parsing models are designed for single-domain systems to answer complex questions. ATIS \citep{price-1990-evaluation, dahl-etal-1994-expanding} contains manually annotated questions for the flight-booking task. GeoQuery \citep{zelle1996geoquery} contains manually annotated questions about US geography. \citet{popescu2003geoquery, giordani-moschitti-2012-translating, iyer-etal-2017-learning} convert GeoQuery into the SQL version. Restaurants \citep{tang-mooney-2000-automated, popescu2003geoquery} is a dataset including questions about restaurants and their food types etc. Scholar \citep{iyer-etal-2017-learning} includes questions about academic publications and corresponding automatically generated SQL queries. Academic \citep{li2014academic} enumerates all query logics supported by the Microsoft Academic Search (MAS) website and writes corresponding question utterances. Yelp and IMDB \citep{yaghmazadeh2017yelp} consists of questions about the Yelp website and the Internet Movie Database. Advising \citep{finegan-dollak-etal-2018-improving} consists of questions about the course information database at the University of Michigan along with artificial data records.

Single-domain text-to-SQL datasets contain only one database. Although text-to-SQL models trained with single-domain datasets are applied in corresponding specific domains, different systems with the same domain but different backgrounds have diverse databases, which means that models should have the generalization ability to be transferred among different systems. Existing single-domain datasets do not own the feature that requires models to improve cross-schema generalization ability. On the contrary, our cross-schema setup is raised for this issue.

\paragraph{Cross-domain text-to-SQL datasets} Recent researches expect text-to-SQL models~\cite{guo-etal-2019-towards,bogin-etal-2019-representing,zhang-etal-2019-editing} to generalize to unseen databases. Thus cross-domain text-to-SQL datasets are released. \citet{zhongSeq2SQL2017} releases WikiSQL, a dataset of 80,654 manually annotated question/SQL pairs distributed across more than 20k tables from Wikipedia. Although WikiSQL is a large-scale dataset, each database schema merely consists of one table and each SQL query merely consists of SELECT, FROM, WHERE clauses. \citet{yu-etal-2018-spider} releases Spider, a large-scale complex cross-domain text-to-SQL dataset. Comparing with previous datasets, Spider owns much more complex databases for various domains and complex SQL queries with advanced SQL clauses and nested SQL structures. \citet{wang-etal-2020-dusql} releases DuSQL, yet another large-scale cross-domain text-to-SQL dataset but in Chinese. Having similar form with Spider, DuSQL has become a popular Chinese text-to-SQL dataset. There are also some conversational cross-domain text-to-SQL datasets, including SParC~\cite{yu-etal-2019-sparc}, CoSQL~\cite{yu-etal-2019-cosql}, CHASE~\cite{guo-etal-2021-chase}, DIR~\cite{li2023dir} etc.

Although our cross-schema dataset owns more than one databases, it is different from cross-domain datasets. It concentrates on model generalization ability across different databases which share the similar structure since they are in the same domain.

\section{Dataset Collection}

In this section, we introduce our method of constructing the medical dataset CSS in detail. The dataset construction method mainly consists of five steps: 1) initial databases creation, 2) question/SQL templates creation, 3) values filling, 4) questions rewriting, and 5) database schema extension.

\begin{figure}[htbp]
    \centering
    \includegraphics[width=0.49\textwidth]{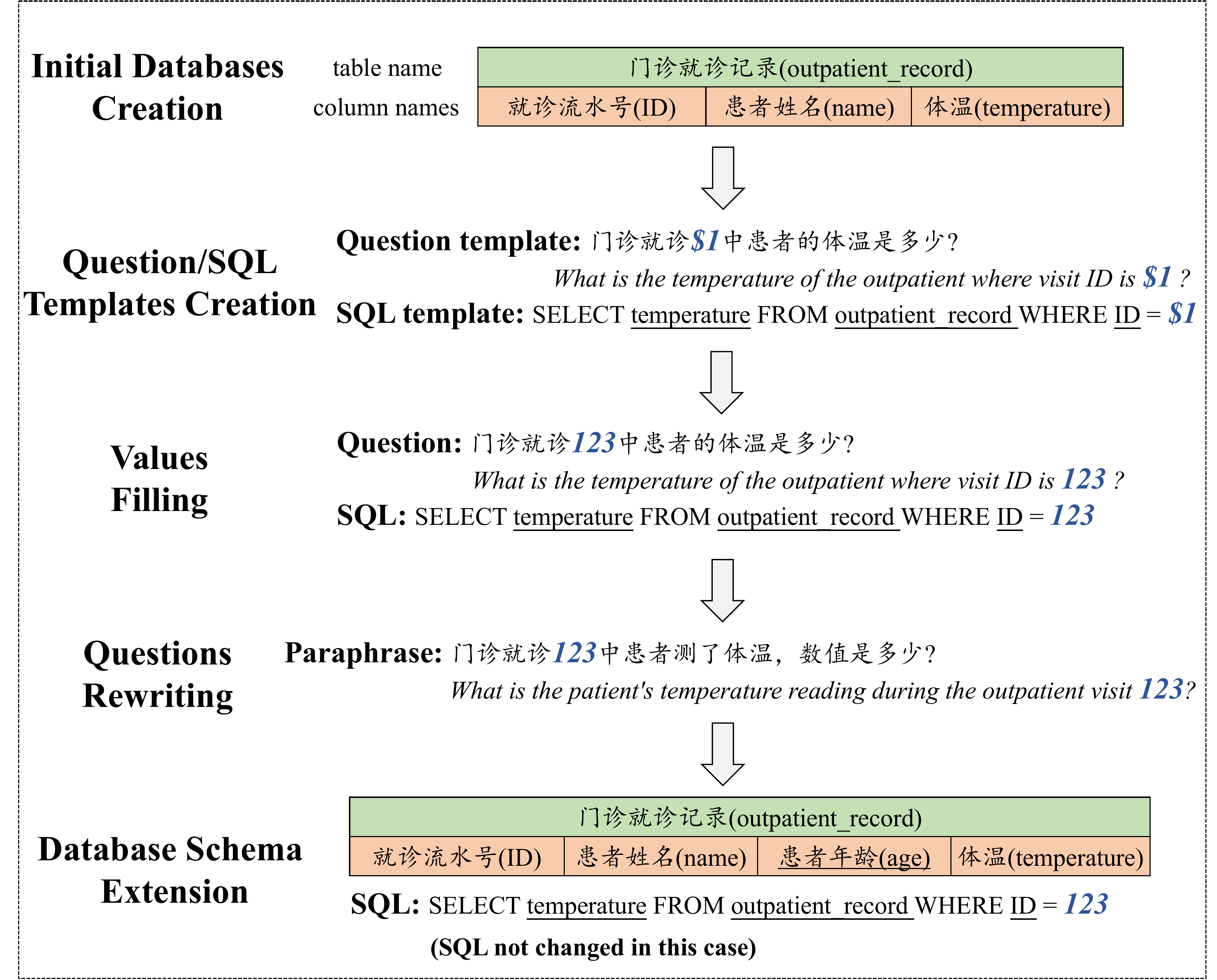}
    \caption{Overview of the dataset collection process.}
    \label{fig:workflow}
\end{figure}

We discuss five steps of constructing the dataset in Section \ref{subsec:initial}-\ref{subsec:extension} respectively. Figure \ref{fig:workflow} shows the overview of the complete process.

\subsection{Initial Databases Creation}
\label{subsec:initial}

To construct the dataset, the first step is to create initial databases. We collect two databases from the real world scenario, i.e. the insurance database and the medical database. The insurance database mainly stores medical consumption records of many different patients. The medical database mainly stores records of medical diagnostic and examination results.

It is obvious that records data in medical databases are usually sensitive, since the issue of patients privacy is involved in these data. It is not feasible to use data from the real world directly in our dataset. To protect privacy of users involved in the medical system, we generate database cell-values with certain rules and ensure that generated data are reasonable.

\subsection{Question/SQL Templates Creation}

Creating abundant and diverse question/SQL templates is an important step for constructing the dataset, which influences the quality of the generated dataset a lot. A question/SQL template can be regarded as an example of the dataset, which consists of a question template and a SQL query template answering the question. The only difference between the question/SQL template and the real dataset example is that values carrying information (e.g. ID, name, time) in the question/SQL template are replaced with special tokens. In the subsequent steps, values can be generated and filled into corresponding question/SQL templates with certain rules, which means that all question/SQL templates can be transformed into real dataset examples eventually.

In general, we use three methods to create various question/SQL templates. Firstly, given medical databases, we enumerate all columns and attempt to raise a question for each column as far as possible. Sometimes we put several columns with close lexical relations into one question/SQL template, since the diversity of the SELECT clause can get increased. It is obvious that question/SQL templates written by this method are relatively simple.

Secondly, we raise a few medical query scenarios and create question/SQL templates based on them. In the real world, different people with different occupations and social roles will ask different types of questions. For instance, patients may care their medical consumption records and doctors may care medical examination results. Based on different real-world scenarios, we can raise various questions that meet needs of people with different social roles (e.g. doctor, patient). Furthermore, these question/SQL templates are usually more challenge since their SQL skeletons are usually more complex and diverse.

Thirdly, we add question/SQL templates which include SQL keywords and SQL skeletons that never occur in previous templates. We count occurrence frequencies for all SQL grammar rules and SQL skeletons that occur in dataset examples. Referring to statistical results, we create questions and corresponding SQL queries which consist of SQL grammar rules that occur in few dataset examples. Detailed statistical results are shown in Section \ref{sec:sql_statistics}. By creating question/SQL templates with this method, the SQL diversity of the dataset can get improved.

We eventually raise 434 different question/SQL templates totally. All these templates will get processed in subsequent steps.

\subsection{Values Filling}

In order to generate real dataset examples from question/SQL templates, values should be generated and filled into all templates. Different types of values are replaced with different special tokens in question/SQL templates. In this step, we use certain rules to generate random values for various special tokens.

Concretely, special tokens indicating number or time are filled with reasonable and suitable random values. Special tokens indicating ID (e.g. person ID, hospital ID) are filled with random strings, which consist of numbers and letters. Other special tokens basically indicate specialized and professional words like disease names. To generate these values, we firstly collect sufficient disease names, medicine names, medical test names, etc. Then these special tokens are filled with values chosen at random from corresponding candidate value lists.

Actually one unique question/SQL template can be used to generate several different dataset examples, since the template can be completed with various random values. We generate 10 dataset examples for each question/SQL template. Consequently there are totally 4,340 question/SQL pairs which are directly generated from 434 question/SQL templates.

\subsection{Questions Rewriting}

Although 4,340 question/SQL pairs directly generated from templates can already be used to train and test text-to-SQL models, they cannot be directly added into the eventual medical dataset. Question sentences generated from question templates are usually unnatural. Moreover, 10 question sentences generated from the same one question template share the same sentence pattern. which means lack of natural language diversity.

To tackle the issue of language naturalness and diversity, we recruit annotators to rewrite dataset examples. All questions directly derived from question templates are rewritten by annotators. In this process, lexical and syntactic patterns of question sentences get changed, which leads to improvement of natural language diversity of the dataset.

To ensure the diversity of rewritten question sentences, we design a specific metric to evaluate the rewriting quality. We recruit two groups of annotators and request them to rewrite question sentences with metric scores as high as possible. Finally we merge two rewriting results from different annotating groups with some rules and acquire all rewritten questions. Detailed explanation of the metric is shown in Appendix \ref{app:rewrite}.

The correctness of rewritten questions is also an important issue. We use the automatic method to examine rewritten questions and make sure that key information are always maintained after the rewriting process.

\textbf{Payment.} All annotators were paid based on their annotations. Annotators would get paid 0.58 RMB for each annotation example.

\subsection{Database Schema Extension}
\label{subsec:extension}

Database schema extension is a key feature of CSS. Text-to-SQL models with good performance should have the ability to be used in various medical systems. In the real world application, different medical systems may use different databases. However, these databases may share the similar structure, since all of them are designed for the medical domain. Consequently, we believe that cross-schema generalization ability for text-to-SQL models is significant and add this challenge task in CSS.

CSS originally contains 2 databases. Based on them, we follow \citet{li2023structural} and create 19 new databases. Firstly for two tables linked with foreign keys, we create a new relation table between the original two tables and create new foreign keys respectively pointing to them. Secondly for two tables linked with foreign keys, we merge them by putting their columns together in a merged table. Thirdly for a table with a special column which only contains a few different kinds of values (e.g. gender), we split the table into several tables according to those limited values.

\begin{figure}[htbp]
    \centering
    \includegraphics[width=0.48\textwidth]{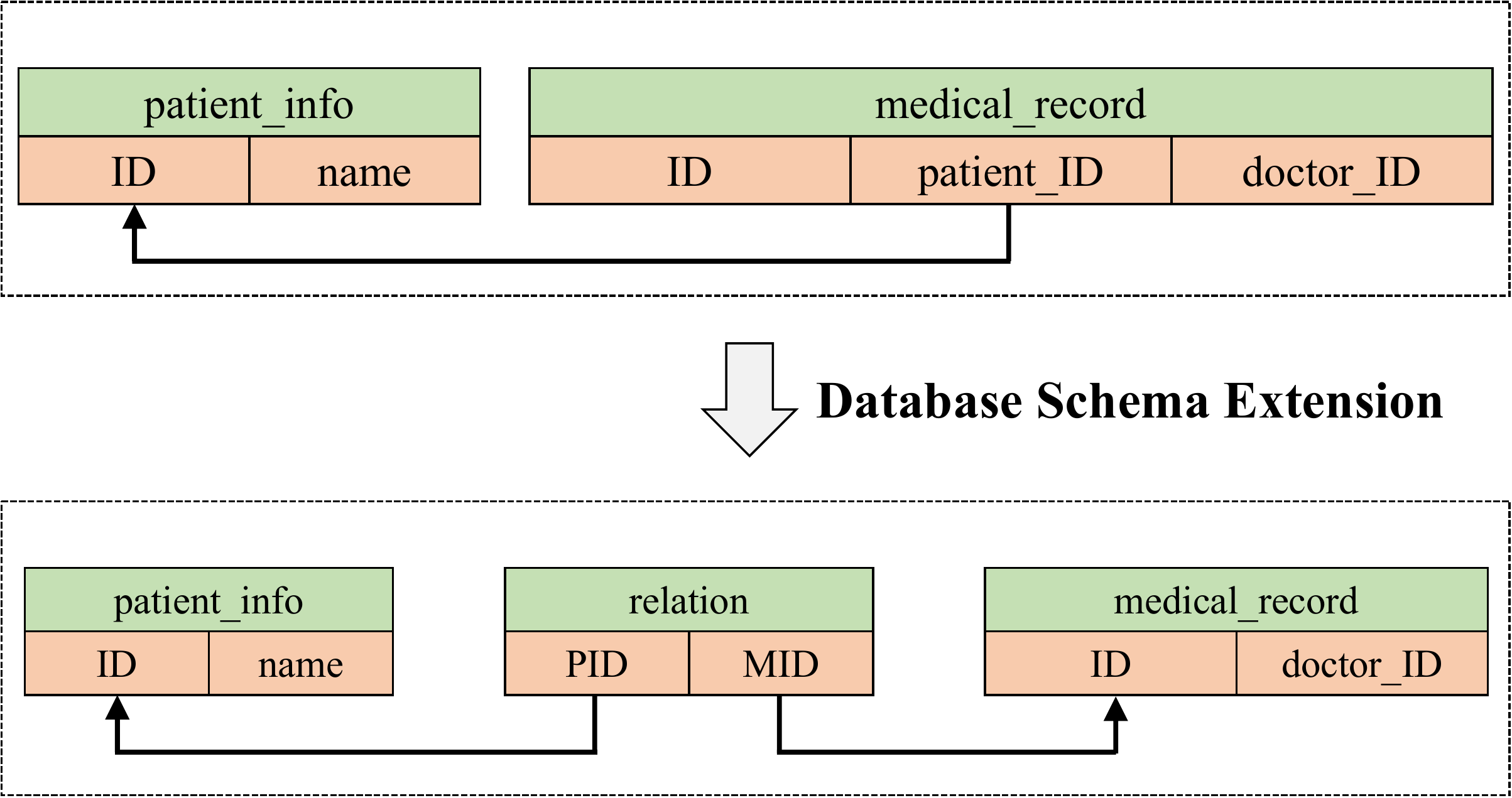}
    \caption{An instance of database schema extension.}
    \label{fig:extend}
\end{figure}

After creating databases, CSS acquires 19 new databases and 29,280 new dataset examples. Therefore, CSS totally contains 33,620 question/SQL pairs across 21 databases.

\section{Dataset Statistics and Comparison}

In this section, we list some statistical information of CSS and existing datasets and do comparison. We mainly discuss scale statistics and SQL statistics with various datasets, including single-domain datasets, cross-domain datasets and CSS.

\subsection{Scale Statistics}

\begin{table*}
\centering
\begin{tabular}{c|ccccccc}
\hline

\hline
\textbf{Dataset} & \textbf{Language} & \textbf{Examples} & \textbf{DBs} & \textbf{Avg T/DB} & \textbf{Avg C/T} & \textbf{Avg P/T} & \textbf{Avg F/T} \\
\hline

\hline
ATIS & English & 19,201 & 1 & 25 & 5.24 & 0.16 & 1.56 \\
GeoQuery & English & 920 & 1 & 8 & 3.88 & 1.75 & 1.12 \\
Restaurants & English & 378 & 1 & 3 & 4.00 & 1.00 & 1.33 \\
Scholar & English & 1,858 & 1 & 12 & 2.33 & 0.58 & 0.75 \\
Academic & English & 200 & 1 & 15 & 2.80 & 0.47 & 0.00 \\
Yelp & English & 141 & 1 & 7 & 5.43 & 1.00 & 0.00 \\
IMDB & English & 147 & 1 & 16 & 4.06 & 1.00 & 0.19 \\
Advising & English & 4,744 & 1 & 18 & 6.89 & 1.39 & 5.39 \\
\hline
WikiSQL & English & 80,654 & 26,531 & 1.00 & 6.34 & 0.00 & 0.00 \\
Spider & English & 9,693 & 166 & 5.28 & 5.14 & 0.89 & 0.91 \\
DuSQL & Chinese & 25,003 & 208 & 4.04 & 5.29 & 0.51 & 0.71 \\
\hline
CSS & Chinese & 33,620 & 21 & 5.62 & 28.49 & 1.68 & 1.65 \\
\hline

\hline
\end{tabular}
\caption{Scale statistics of existing datasets. "Avg T/DB" represents the average number of tables per database schema. "Avg C/T" represents the average number of columns per table. "Avg P/T" represents the average number of columns in the composite primary key per table. "Avg F/T" represents the average number of foreign keys per table.}
\label{tab:scale_statistics}
\end{table*}

Table \ref{tab:scale_statistics} shows scale statistics of existing datasets, including single-domain datasets, cross-domain datasets, and the medical dataset CSS. For single-domain datasets listed in the table and WikiSQL, we use the standardized version from \citet{finegan-dollak-etal-2018-improving}. CSS contains 33,620 examples generated from scratch across 21 databases. Comparing with previous single-domain datasets, CSS has the largest scale and various databases. We extend original databases with several certain rules. Therefore, CSS can help text-to-SQL models generalize to different medical systems, where databases are different but share the similar structure.

Databases in CSS have a great number of columns, composite primary keys, and foreign keys, which indicates that databases in CSS commonly possess complex structures. This is also a challenge feature of CSS. It requires models to find out effective information from complex database structures.

\subsection{SQL Statistics}
\label{sec:sql_statistics}

First of all, we clarify the concept named SQL skeleton. For a certain SQL query, it is feasible to remove detailed schema items and values from the SQL query. Concretely, we replace tables used in the SQL query with the special token "tab". Columns and values are processed with the similar method. Columns are replaced with the special token "col" and values are replaced with the special token "value". Then the result is defined as the SQL skeleton, which retains the basic structure of the original SQL query.

Table \ref{tab:sql_statistics} shows SQL statistics of existing datasets. CSS totally possesses 562 different SQL skeletons, which is comparable with ATIS and surpasses other single-domain datasets. Note that SQL queries in CSS are commonly very long. The average and maximum number of SQL query tokens are 55.41 and 243 respectively, which has surpassed almost all existing datasets except ATIS. The statistical result indicates that SQL queries in CSS are diverse and complex. This is still a challenge for text-to-SQL models.

\section{Tasks and Models}

\subsection{Dataset Splitting}

We provide three methods to split the dataset into train/dev/test sets. Different dataset splitting methods correspond to different tasks and raise different challenges for models. For the first method, 4,340 original dataset examples are shuffled at random and then are split with the ratio 0.8/0.1/0.1. This sub-task is an ordinary text-to-SQL task setting and requires models to generalize well on natural language.

For the second method, 434 question/SQL templates are shuffled at random and then are split with the ratio 0.8/0.1/0.1. Then 4,340 original question/SQL pairs fall into corresponding dataset subsets. Comparing with other dataset splitting methods, larger language gap and SQL gap exist among train/dev/test sets, since different question/SQL templates generally express different meanings. Models are required to have the stronger SQL pattern generalization ability under this sub-task.

For the third method, we add extended dataset examples and split all 33,620 examples according to their databases. All databases are split with the ratio 0.6/0.2/0.2. No overlap of databases exists in train/dev/test sets. This dataset splitting method provides a challenge task, which requires models to possess the stronger generalization ability across diverse databases sharing similar structures.

\subsection{Syntactic Role Prediction}
\label{subsec:srp}

How to improve the cross-schema generalization ability of text-to-SQL models is a key challenge raised in CSS. In this section, we introduce our simple method to tackle the issue of model generalization ability across different databases.

The text-to-SQL model LGESQL \citep{cao-etal-2021-lgesql} add an auxiliary task named graph pruning in order to improve the model performance. Given the natural language question and the database schema, the model is required to predict whether each schema item occurs in the SQL query. Following \citet{cao-etal-2021-lgesql}, we raise a similar auxiliary task named syntactic role prediction (SRP). Under this task, the model is required to predict in which SQL clause each question token occurs.

\begin{table}
\centering
\begin{tabular}{c|ccccccc}
\hline

\hline
\textbf{Dataset} & \textbf{\# SQL} & \textbf{Avg Len} & \textbf{Max Len} \\
\hline

\hline
ATIS & 828 & 97.96 & 474 \\
GeoQuery & 120 & 26.08 & 92 \\
Restaurants & 12 & 29.22 & 61 \\
Scholar & 158 & 37.07 & 65 \\
Academic & 76 & 36.30 & 116 \\
Yelp & 62 & 28.92 & 56 \\
IMDB & 30 & 27.48 & 55 \\
Advising & 169 & 47.49 & 169 \\
\hline
WikiSQL & 39 & 12.48 & 23 \\
Spider & 1,116 & 17.99 & 87 \\
DuSQL & 2,323 & 20.23 & 37 \\
\hline
CSS & 562 & 55.41 & 243 \\
\hline

\hline
\end{tabular}
\caption{SQL statistics of existing datasets. "\# SQL" represents the number of SQL skeletons. "Avg Len" represents the average number of tokens in one SQL query. "Max Len" represents the maximum number of tokens in one SQL query.}
\label{tab:sql_statistics}
\end{table}

\begin{figure}[htbp]
    \centering
    \includegraphics[width=0.48\textwidth]{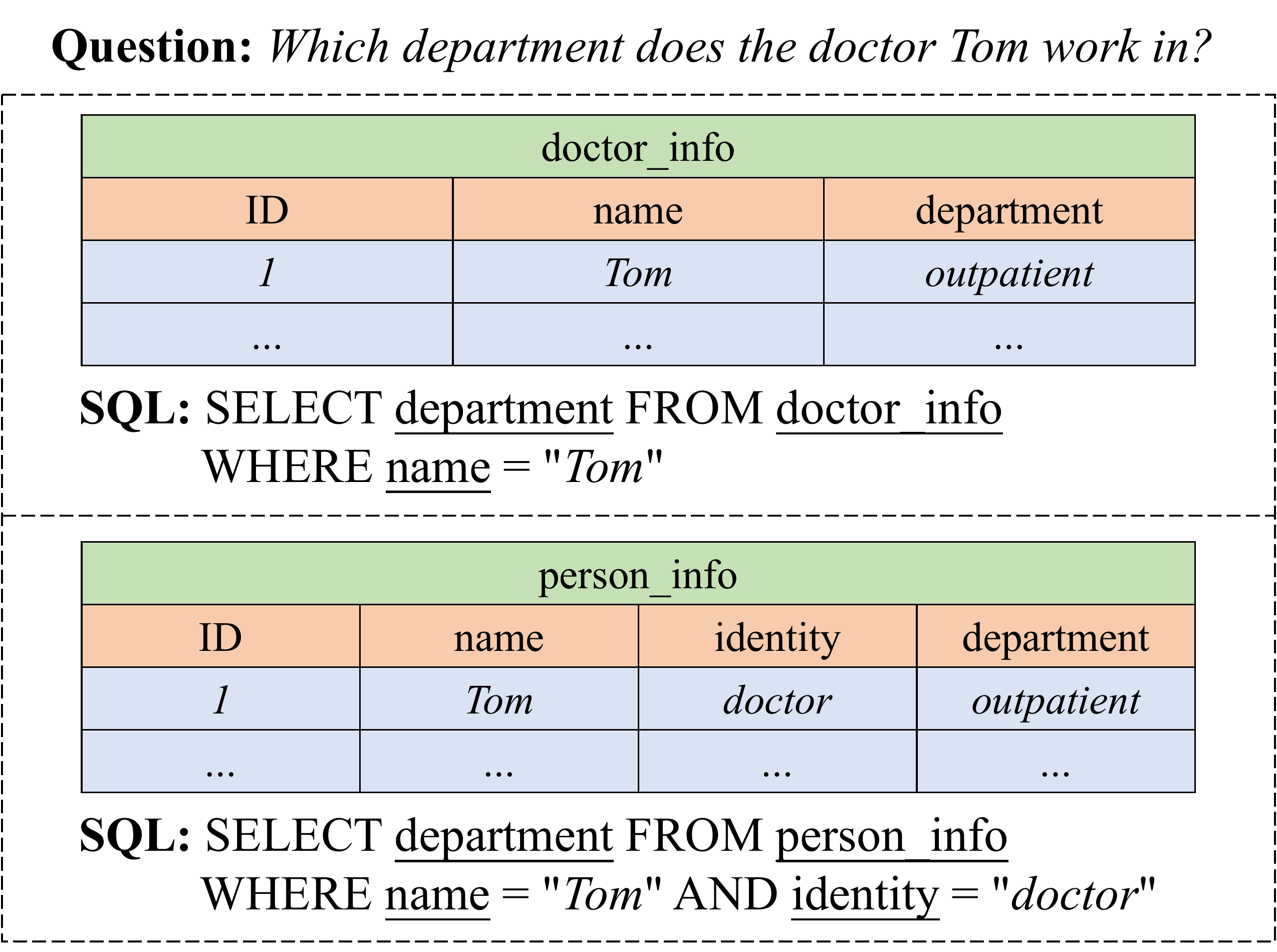}
    \caption{Given the question, the corresponding SQL query differs among various but similar databases.}
    \label{fig:srp}
\end{figure}

The SQL query structure may change as the database schema changes. Figure \ref{fig:srp} shows an instance, where two databases share the similar structure but the key information "doctor" in the question are used in the FROM clause and the WHERE clause respectively. We hypothesize that model with strong cross-schema generalization ability should distinguish syntactic roles of every question tokens under different databases.

Concretely, according to the text-to-SQL model LGESQL, the model input is a graph $G = (V, E)$ constructed with the given question and the database schema. Graph nodes $V$ include question tokens and schema items (i.e. tables and columns) and graph edges $E$ indicate relations among them. The model encodes each node $i$ into an embedding vector $\mathbf{x}_i$. Then the context vector $\tilde{\mathbf{x}}_i$ for each node $i$ can be computed with multi-head attention.
\begin{align*}
\alpha_{i j}^h & = \mathrm{softmax}_{j \in \mathcal{N}_i} \frac{(\mathbf{x}_i \mathbf{W}_q^h) (\mathbf{x}_j \mathbf{W}_k^h)^\mathrm{T}}{\sqrt{d / H}}, \\
\tilde{\mathbf{x}}_i & = (\mathrm{concat}_{h = 1}^H \sum_{j \in \mathcal{N}_i} \alpha_{i j}^h \mathbf{x}_j \mathbf{W}_v^h) \mathbf{W}_o,
\end{align*}
where $d$ is the dimension of embedding vectors, $H$ is the number of heads, $\mathcal{N}_i$ is the neighborhood of the node $i$, and $\mathbf{W}_q^h, \mathbf{W}_k^h, \mathbf{W}_v^h \in \mathbb{R}^{d \times d / H}, \mathbf{W}_o \in \mathbb{R}^{d \times d}$ are network parameters.

For each question node $q_i$, the model can predict in which SQL clause it occurs with $\mathbf{x}_{q_i}$ and $\tilde{\mathbf{x}}_{q_i}$. Specifically we divide the SQL query into 16 different parts, which are discussed in detail in Appendix \ref{app:srp}. Thus the auxiliary task is a binary classification task for each question token and each SQL part.
$$
P(\mathbf{y}_{q_i} | \mathbf{x}_{q_i}, \tilde{\mathbf{x}}_{q_i}) = \sigma([\mathbf{x}_{q_i}; \tilde{\mathbf{x}}_{q_i}] \mathbf{W} + \mathbf{b}),
$$
where $W \in \mathbb{R}^{2 d \times 16}, b \in \mathbb{R}^{1 \times 16}$ are network parameters and $\mathbf{y}_{q_i}$ is the probability vector. The ground truth $y_{q_i, j}^g$ is 1 when the question token $q_i$ occurs in the $j$-th SQL part. The training object is
\begin{align*}
\mathcal{L} & = - \sum_{q_i} \sum_j [y_{q_i, j}^g \log P(y_{q_i, j} | \mathbf{x}_{q_i}, \tilde{\mathbf{x}}_{q_i}) \\
& + (1 - y_{q_i, j}^g) \log(1 - P(y_{q_i, j} | \mathbf{x}_{q_i}, \tilde{\mathbf{x}}_{q_i})].
\end{align*}

The syntactic role prediction task is combined with the main task in a multitasking way. In addition, SRP can also be added into the RATSQL model directly, since RATSQL and LGESQL both encode graph nodes into embedding vectors and SRP only takes these vectors as the input.

\section{Experiments}

\subsection{Experiment Setup}

\paragraph{Baseline approaches} We adopt three competitive text-to-SQL models as the baseline approaches, i.e. RATSQL \citep{wang-etal-2020-rat}, LGESQL \citep{cao-etal-2021-lgesql}, and PICARD \citep{scholak-etal-2021-picard}. RATSQL and LGESQL process given information with graph encoding and decode the abstract syntax tree (AST) of the result SQL query. PICARD is a sequence-to-sequence approach and is different from the other two approaches.

RATSQL constructs a graph with question tokens and schema items (i.e. tables and columns) and encodes the graph with the relation-aware self-attention mechanism. With the unified framework, RATSQL can easily establish and handle relations among graph nodes and then encode elements with various categories jointly.

Comparing with RATSQL, LGESQL improves the model performance by utilizing the line graph. LGESQL pays more attention to the topological structure of graph edges and distinguishes local and non-local relations for graph nodes. Besides the original graph used in RATSQL, LGESQL also constructs the corresponding line graph, since the line graph can help facilitate propagating encoding messages among nodes and edges.

Different from RATSQL and LGESQL, PICARD is a sequence-to-sequence model. Nowadays large pretrained language models have possessed the strong ability for handling and processing natural language with unconstrained output space. However, SQL is a formal language with strict grammar rules. Invalid SQL queries are very likely to be generated if pretrained models are directly finetuned with text-to-SQL datasets. PICARD provides an approach, which can help reject invalid tokens during each decoding step and generate sequences in the constrained output space.

For each baseline model, we use pretrained language models (PLMs) within the encoding module. In our experiments, the PLM \texttt{longformer-chinese-base-4096} is applied in RATSQL and LGESQL and the PLM \texttt{mbart-large-50} is applied in PICARD.

\paragraph{Evaluation metrics} There are several metrics to evaluate text-to-SQL model performances, including exact match and execution accuracy etc. The exact match metric requires the predicted SQL query to be equivalent to the gold SQL query. The execution accuracy metric requires the execution result of the predicted SQL query to be correct.

We mainly use the exact match (EM) metric in our experiments. Concretely, we present model performances with (w) and without (w/o) value evaluation respectively.

\subsection{Results and Analysis}

\begin{table}
\centering
\begin{tabular}{c|cc|cc}
\hline

\hline
\multirow{2}*{\textbf{Model}} & \multicolumn{2}{c}{\textbf{Dev}} & \multicolumn{2}{|c}{\textbf{Test}} \\
\cline{2-5}
~ & \textbf{w/o} & \textbf{w} & \textbf{w/o} & \textbf{w} \\
\hline

\hline
RATSQL & 90.2 & 81.1 & 89.0 & 79.1 \\
LGESQL & 91.7 & 82.2 & 90.8 & 81.1 \\
PICARD & 93.8 & 53.7 & 70.3 & 58.3 \\
\hline

\hline
\end{tabular}
\caption{Model performances under dataset splitting method according to examples.}
\label{tab:example}
\end{table}

According to 3 different dataset splitting methods, we test baseline models under 3 sub-task settings. Table \ref{tab:example} shows model performances under dataset splitting method according to examples. LGESQL achieves the best performance under this sub-task, i.e. 90.8\% EM(w/o) accuracy and 81.1\% EM(w) accuracy on the test set. This indicates that existing text-to-SQL parsing models have had the ability to perform very well if all databases and possible SQL structures have appeared in the train set. Models merely need to generalize on natural language, which is simple when utilizing strong PLMs.

Table \ref{tab:template} shows model performances under the template-splitting sub-task. Comparing with the previous sub-task, performances of three baseline models decrease a lot. Although RATSQL achieves the best performance under this sub-task, the EM(w/o) accuracy and the EM(w) accuracy on the test set are only 58.9\% and 53.0\% respectively. Question/SQL templates in dev/test sets do not appear in the train set. Thus models have to predict unseen SQL patterns when testing. The experiment result indicates that there is still a large room for the improvement of model generalization ability across SQL patterns. We believe that CSS can also help facilitate researches on improving model ability of predicting unseen SQL patterns.

\begin{table}[htbp]
\centering
\begin{tabular}{p{\columnwidth}}
\hline

\hline
\textbf{Q:} \begin{CJK*}{UTF8}{gkai}列出水天干这位患者在医院7539997住院的就诊记录里入院科室名字含有\textcolor[RGB]{0,112,192}{耳鼻喉科}的记录\end{CJK*} \\
\textbf{Q:} {\it List the records of patient Tiangan Shui admitted to hospital 7539997, including the records with the department name containing \textcolor[RGB]{0,112,192}{Otolaryngology}.} \\
\textbf{Gold:} {\small SELECT * FROM person\_info JOIN hz\_info JOIN zyjzjlb WHERE person\_info.XM = "\begin{CJK*}{UTF8}{gkai}水天干\end{CJK*}" AND hz\_info.YLJGDM = "{\it 7539997}" AND zyjzjlb.JZKSMC LIKE "\%\begin{CJK*}{UTF8} {gkai}\textcolor[RGB]{0,112,192}{\textbf{耳鼻喉科}}\end{CJK*}\%"} \\
\textbf{Pred:} {\small SELECT * FROM person\_info JOIN hz\_info JOIN zyjzjlb WHERE person\_info.XM = "\begin{CJK*}{UTF8}{gkai}水天干\end{CJK*}" AND hz\_info.YLJGDM = "{\it 7539997}" AND zyjzjlb.JZKSMC LIKE "\%\begin{CJK*}{UTF8}{gkai}\textcolor[RGB]{176,23,31}{耳鼻炎}\end{CJK*}\%"} \\
\hdashline
\textbf{Q:} \begin{CJK*}{UTF8}{gkai}从01年1月31日一直到09年8月12日内患者80476579被开出\textcolor[RGB]{0,112,192}{盐酸多奈哌齐片(薄膜)}的总次数一共有多少？\end{CJK*} \\
\textbf{Q:} {\it How many times has patient 80476579 been prescribed \textcolor[RGB]{0,112,192}{donepezil hydrochloride tablets (thin film)} from 2001-01-31 to 2009-08-12?} \\
\textbf{Gold:} {\small SELECT COUNT(*) FROM t\_kc21 JOIN t\_kc22 WHERE t\_kc21.PERSON\_ID == "{\it 80476579}" AND t\_kc22.STA\_DATE BETWEEN "{\it 2001-01-31}" AND "{\it 2009-08-12}" AND t\_kc22.SOC\_SRT\_DIRE\_NM == "\begin{CJK*}{UTF8}{gkai}\textcolor[RGB]{0,112,192}{盐酸多奈哌齐片(薄膜)}\end{CJK*}"} \\
\textbf{Pred:} {\small SELECT COUNT(*) FROM t\_kc21 JOIN t\_kc22 WHERE t\_kc21.PERSON\_ID == "{\it 80476579}" AND t\_kc22.STA\_DATE BETWEEN "{\it 2001-01-31}" AND "{\it 2009-08-12}" AND t\_kc22.SOC\_SRT\_DIRE\_NM == "\begin{CJK*}{UTF8}{gkai}\textcolor[RGB]{176,23,31}{盐酸多奈}\end{CJK*}"} \\
\hline

\hline
\end{tabular}
\caption{Case study for the PICARD model when predicting values. FROM conditions are omitted for clarity.}
\label{tab:case}
\end{table}

Note that as a sequence-to-sequence approach, PICARD cannot perform as well as the two AST-based approaches (RATSQL and LGESQL) in the template-splitting sub-task. There is a room of model performances between PICARD and AST-based approaches, especially when values in SQL queries are concerned in evaluation. Table \ref{tab:case} shows two instances from the test set in the template-splitting sub-task, where the PICARD model successfully generates the structure of the SQL query but predicts the wrong value. As shown in Table \ref{tab:sql_statistics}, SQL queries in CSS are commonly very long and complex, which leads to great difficulty for PICARD decoding. The decoding error would accumulate as the decoding step increases. According to our statistical results, during the decoding process of AST-based approaches, the average number of AST nodes is 56.95. Although the average number of tokens in the SQL query is 55.41, PLM used in PICARD would split tokens into many sub-words. Consequently, decoding steps of PICARD is actually much more than AST-based approaches. Furthermore, table and column names in CSS are commonly consisted of unnatural tokens, which improves the decoding difficulty of PICARD a lot.

\begin{table}
\centering
\begin{tabular}{c|cc|cc}
\hline

\hline
\multirow{2}*{\textbf{Model}} & \multicolumn{2}{c}{\textbf{Dev}} & \multicolumn{2}{|c}{\textbf{Test}} \\
\cline{2-5}
~ & \textbf{w/o} & \textbf{w} & \textbf{w/o} & \textbf{w} \\
\hline

\hline
RATSQL & 60.5 & 55.2 & 58.9 & 53.0 \\
LGESQL & 59.5 & 54.4 & 58.5 & 52.8 \\
PICARD & 52.6 & 40.9 & 49.8 & 38.6 \\
\hline

\hline
\end{tabular}
\caption{Model performances under dataset splitting method according to templates.}
\label{tab:template}
\end{table}

\begin{table}
\centering
\begin{tabular}{c|cc|cc}
\hline

\hline
\multirow{2}*{\textbf{Model}} & \multicolumn{2}{c}{\textbf{Dev}} & \multicolumn{2}{|c}{\textbf{Test}} \\
\cline{2-5}
~ & \textbf{w/o} & \textbf{w} & \textbf{w/o} & \textbf{w} \\
\hline

\hline
RATSQL & 36.6 & 35.7 & 43.4 & 42.0 \\
RATSQL + SRP & 38.3 & 37.4 & 47.2 & 45.3 \\
\hline

\hline
\end{tabular}
\caption{Model performances under dataset splitting method according to databases. "SRP" represents the auxiliary task named syntactic role prediction.}
\label{tab:schema}
\end{table}

Table \ref{tab:schema} shows model performances under dataset splitting method according to different databases. Under this sub-task, we use RATSQL as the baseline model and attempt to add the auxiliary task SRP, expecting to improve the model performance across different databases. The experiment result shows that the model performance increases about 1.7\% on the dev set and increases about 3.3\%-3.8\% on the test set when SRP is applied into RATSQL. This proves that SRP can help improve the cross-schema generalization ability of the model when using SRP as a simple baseline method.

\begin{figure}[htbp]
    \centering
    \includegraphics[width=0.48\textwidth]{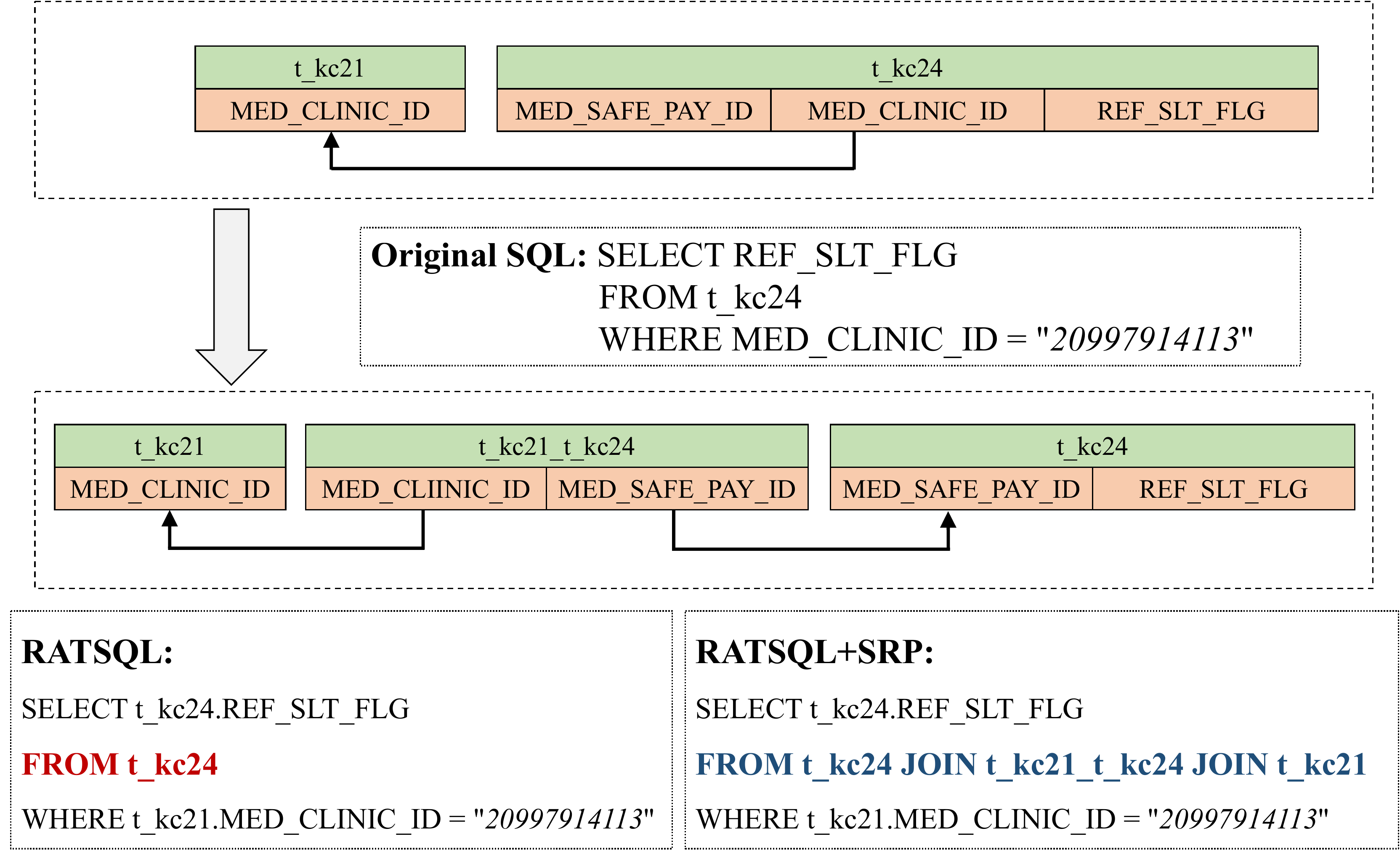}
    \caption{Case study for SRP. JOIN conditions in the FROM clause are omitted for brevity.}
    \label{fig:case}
\end{figure}

Figure \ref{fig:case} is an instance from the test set, where RATSQL predicts the wrong SQL but RATSQL with SRP predicts the correct result. After database schema extension, a new relation table is created. However, RATSQL does not understand the change and misses the relation table in the FROM clause. On the contrary, the auxiliary task SRP helps the model utilize the relation table and eventually predict the correct SQL.

\section{Conclusion}

This paper presents CSS, a large-scale cross-schema Chinese text-to-SQL dataset designed for the medical domain. We illustrate the detailed process of dataset construction and also present statistical information comparing with existing datasets. We raise a challenge task in CSS, which requires models to generalize across various databases but in the same domain. To tackle the above task, we designed a baseline method named syntactic role prediction as an auxiliary task for model training. We conduct benchmark experiments with three competitive baseline models and prove that future researches on CSS is valuable.

\section*{Limitations}

We raise a new challenge task in our medical dataset CCS. Comparing with existing datasets, CCS requires text-to-SQL models to generalize to different databases with the similar structure in the same domain. To tackle this problem, we provide a baseline method named syntactic role prediction, which is an auxiliary task and can be combined with the main task in a multitasking way. Our experiments prove that SRP can help improve the cross-schema generalization ability of models. However, the improvement is not that large. How to generalize models across different databases sharing the similar structure is still a challenge issue. We expect that future works can solve this difficult problem.

\section*{Ethics Statement}

We collect two original medical databases from the real world. However, cell-values in medical databases are commonly sensitive, since the information of patients and doctors are involved in these values. Thus we only retain the database schema and generate sufficient cell-values with certain rules. We ensure that generated values are reasonable and that privacy of medical system users can get protected.

\section*{Acknowledgments}

We thank Xiaowen Li, Kunrui Zhu and Ruihui Zhao from Tencent Jarvis Lab for providing necessary initial data. We also thank all the anonymous reviewers for their thoughtful comments. This work has been supported by the China NSFC Project (No.62106142 and No.62120106006), Shanghai Municipal Science and Technology Major Project (2021SHZDZX0102), CCF-Tencent Open Fund
and Startup Fund for Youngman Research at SJTU (SFYR at SJTU).

\bibliography{anthology,custom}
\bibliographystyle{acl_natbib}

\appendix

\section{Rewriting Metric}
\label{app:rewrite}

First of all, we define the rewriting ratio (RR) between two different sentences $s_1$ and $s_2$, i.e.
$$
RR(s_1, s_2) = \frac{\mathrm{EditDistance}(s_1, s_2)}{|s_1| + |s_2|},
$$
where $\mathrm{EditDistance}(s_1, s_2)$ represents the edit distance between $s_1$ and $s_2$. Assume that $s_{i, 1}, s_{i, 2}, \cdots, s_{i, 10}$ are ten rewritten question sentences derived from the same question/SQL template $i$. In order to improve the language diversity, we expect ten rewritten sentences to differ from each other. Thus we request annotators to maximize
$$
\frac{1}{N} \sum_{i = 1}^N \frac{1}{55} \sum_{1 \le j < k \le 10} RR(s_{i, j}, s_{i, k}),
$$
when rewriting, where $N$ is the number of question/SQL templates.

When merging rewriting results from two groups of annotators, for each example with the original question sentence $s^o$, we need to decide between two rewritten sentences $s_1^r$ and $s_2^r$. Here we choose $s_1^r$ only if
$$
RR(s^o, s_1^r) > RR(s^o, s_2^r).
$$

\section{Syntactic Role Prediction}
\label{app:srp}

\begin{table*}
\centering
\begin{tabular}{c|c}
\hline
\textbf{Name} & \textbf{Description} \\
\hline
NONE & Element is not used in SQL. \\
SELECT & Element is a normal column in SELECT. \\
SELECT\_AGG & Element is a column with an aggregation function in SELECT. \\
SELECT\_NEST & Element appears in SELECT, where SELECT is a nested SQL query. \\
FROM & Element is a normal table in FROM. \\
FROM\_NEST & Element appears in FROM, where FROM is a nested SQL query. \\
WHERE & Element is a normal column in WHERE \\
WHERE\_NEST & Element appears in WHERE, where WHERE is a nested SQL query \\
GROUP & Element is a normal column in GROUP BY. \\
HAVING & Element appears in HAVING. \\
ORDER & Element is a normal column in ORDER BY. \\
ORDER\_AGG & Element is a column with an aggregation funciton in ORDER BY. \\
LIMIT & Element appears in LIMIT. \\
INTERSECT & Element appears in INTERSECT. \\
UNION & Element appears in UNION. \\
EXCEPT & Element appears in EXCEPT. \\
\hline
\end{tabular}
\caption{16 parts of the SQL query.}
\label{tab:srp}
\end{table*}

We divide the SQL query into 16 different parts. Table \ref{tab:srp} shows detailed situations. For each question token $q_i$, we find out all schema items which have schema linking relations with $q_i$. Then for each SQL part, we label that $q_i$ appears in this part if $q_i$ itself or one of those schema items appears in this part.

\end{document}